\documentclass[]{ceurart}
\usepackage[utf8]{inputenc}
\usepackage{csquotes}
\usepackage{booktabs}
\usepackage{tabularx}
\usepackage{amsmath}
\usepackage{amssymb}
\usepackage{graphicx}
\usepackage{caption}
\usepackage{float}
\usepackage{microtype}
\usepackage{tikz}
\usepackage{pgfplots}
\pgfplotsset{compat=1.18}
\usepackage{pgfplotstable}
\usetikzlibrary{positioning,shapes.geometric,arrows.meta,fit,backgrounds,calc}

\begin{document}

\title{Snugi-AI-v2 @ eRisk 2026 Task 2: Early Depression Detection
       via a Learned Stopping Policy with Sustained Confidence Gate}

\author[1]{Yuwen Chiu}[%
    orcid=0009-0008-2859-4679,
    email=ychiu60@gatech.edu
]
\cormark[1]
\address[1]{Georgia Institute of Technology, North Ave NW,
            Atlanta, GA 30332}
\cortext[1]{Corresponding author.}

\begin{abstract}
We describe the Snugi-AI-v2 submission to eRisk 2026 Task~2, the second
edition of contextualized early depression detection from Reddit discussions.
Our central contribution is a learned MLP stopping policy trained to directly
optimize ERDE50, replacing the fixed and tiered threshold strategies used in
all prior eRisk Task~2 submissions. Combined with a sustained confidence gate
that commits only after $N{=}3$ consecutive rounds of high policy confidence,
the system reduces false positives caused by transient emotional posts without
sacrificing recall. The pipeline encodes each discussion thread with a frozen
MentalRoBERTa model, maps the accumulated representation to a depression
probability via an MLP classifier, and delegates the timing decision to the
learned policy. Our best run achieves $F_1 = 0.73$ (Run~1) and $F_{\text{latency}} = 0.70$
(Runs~0 and~3), with a median alert round of~8 out of 500,
completing the full evaluation in 1~hour~26~minutes, the fastest
among all complete-submission teams. We report a systematic ablation across five
runs spanning two encoder variants, four stopping strategies, and three gate values, along
with negative results from GRPO policy training, BDI-II post filtering,
MentalLongformer encoding, and DeBERTa ensembling.
Code: \url{https://github.com/chiuyuwen91/erisk-2026}
\end{abstract}


\conference{CLEF 2026 Working Notes, 21--24 September 2026, Jena, Germany}

\copyrightyear{2026}
\copyrightclause{Copyright for this paper by its authors.
Use permitted under Creative Commons License Attribution 4.0
International (CC BY 4.0).}

\maketitle

\section{Introduction}
Early detection of mental health crises from social media has clear clinical
value: identifying at-risk individuals before a crisis allows timely
intervention. The eRisk lab has standardized this problem since 2017,
introducing ERDE as an evaluation metric that jointly penalizes delayed
true positive alerts and false positive alerts~\cite{losada2016}. The 2026
edition of Task~2 provides the complete Reddit conversation thread for each
target user, including all other participants' posts, a realistic setting
where the clinical relevance of a message often depends on the surrounding
dialogue~\cite{PerezEtAl2026eRiskLNCS}.

Framing this as a classification problem misses something important. A system
does not only need to decide \emph{whether} a user is depressed; it also needs to decide \emph{when} to commit. The ERDE metric makes this explicit: an
alert at round~7 costs far less than an alert at round~70, even if both are
correct. Yet prior eRisk Task~2 systems treat the timing decision as secondary,
using fixed probability thresholds, tiered thresholds by round number, or
the risk window of Sadeque et al.~\cite{sadeque2018}, which requires $n$
consecutive positive predictions before committing. None of these strategies
are \emph{trained} to minimize ERDE50. They encode domain intuitions that
may not transfer across datasets or task editions.

We address this directly with a \emph{learned stopping policy}: a small MLP
that takes a 5-dimensional summary of the probability trajectory as input and
outputs a commitment probability. The policy is trained end-to-end to produce
decisions that minimize ERDE50 on the training data. We combine it with a
sustained confidence gate, requiring the policy to output high
commitment probability for $N$ consecutive rounds before firing; which
filters out transient emotional spikes that cause false alerts. The gate
extends the risk window concept~\cite{sadeque2018} from a hand-tuned rule to
a component of a learned system.

We submitted five runs across two encoders and four stopping strategies,
systematically ablating each design decision. Our best runs achieve
$F_1 = 0.73$ and $F_{\text{latency}} = 0.70$, completing the 500-round
evaluation in 1~hour~26~minutes, the fastest complete submission among 16 teams.

\paragraph{Contributions.}
(1)~A learned stopping policy that directly optimizes ERDE50,
the first such policy in the eRisk Task~2 setting; all prior Task~2 systems use
fixed or heuristic thresholds.
(2)~A sustained confidence gate ($N{=}3$) that extends the risk
window concept to a learned policy, reducing impulsive false alerts.
(3)~A systematic ablation across five runs spanning two encoder variants, four stopping
strategies, and three gate values in a single framework.
(4)~Documented negative results from GRPO~\cite{shao2024grpo} policy training, BDI-II
post filtering, MentalLongformer~\cite{ji2023mentallongformer}, and DeBERTa~\cite{he2021debertav3} ensembling.

\section{Related Work}
Early risk detection on social media has been studied through the eRisk lab
since 2017~\cite{losada2016}. Early systems relied on hand-crafted features
combined with classical classifiers. More recent work uses transformer-based
encoders, with MentalRoBERTa~\cite{ji2022}, a RoBERTa model continued-pretrained
on 13.6~million mental health Reddit sentences, which serves as a strong
general-purpose encoder for this domain.

The question of \emph{when} to alert received direct attention from Sadeque
et al.~\cite{sadeque2018}, who proposed $F_{\text{latency}}$ as a
latency-aware evaluation metric and introduced the \emph{risk window}: a
system must predict positive for $n$ consecutive posts before committing,
preventing impulsive alerts on single emotional spikes. They showed this
technique improves sequential model performance. Our work extends this idea:
instead of applying a risk window to a fixed threshold, we apply a sustained
gate to a \emph{learned} policy that is trained to output the right commitment
probability for each moment in the trajectory.

eRisk 2025 introduced the contextualized variant of the task, providing full
Reddit discussion threads rather than isolated posts~\cite{parapar2025overview}. The
top-performing system, HIT-SCIR, responded to the resulting train/test
distribution gap through LLM-based data augmentation, BDI-II psychiatric
scale-guided post screening, and a hierarchical attention network~\cite{hitscir2025}.
This approach is effective but computationally expensive and dependent on the
specific distribution gap that existed in 2025. In 2026, training and test
data both contain real conversational context, removing the augmentation
motivation~\cite{PerezEtAl2026eRiskLNCS}.

\section{Methodology}
\subsection{Task and Evaluation}

Task~2 follows a sequential protocol. For each target user $u$, the server
releases discussion threads one per round. After each round, the system
submits a binary decision $d_u \in \{0,1\}$ and a confidence score.
Decision~1 is final; decision~0 is reversible. Performance is measured
with ERDE50:
\begin{equation}
  \text{ERDE}_{50}(S) = \frac{1}{|U|} \sum_{u \in U} c(u)
\end{equation}
where the per-user cost $c(u)$ is:
\begin{equation}
  c(u) = \begin{cases}
    c_{\text{FN}} & \text{if depressed, no alert} \\
    \ell_c(k_u) \cdot c_{\text{FN}} & \text{if depressed, alerted at round } k_u \\
    c_{\text{FP}} & \text{if control, alerted} \\
    0 & \text{otherwise}
  \end{cases}
\end{equation}
with latency cost $\ell_c(k) = 1 - \tfrac{1}{1 + e^{k-50}}$,
$c_{\text{FN}} = 1.0$, and $c_{\text{FP}} = 0.1296$. The evaluation also
reports $F_{\text{latency}} = F_1 \times \text{speed}$, where
$\text{speed} = 1 - \text{median}\{\ell_c(k_u)\}$ over true positives.

A key observation is that ERDE50 is \emph{not} minimized by maximizing $F_1$ alone. The latency cost $\ell_c(k)$ grows slowly for the first 50 rounds
and then accelerates. A system that fires at round~7 on average accumulates
far less latency cost than one that fires at round~50, even if both achieve
the same recall. This asymmetry motivates a dedicated \emph{stopping policy}
trained to recognize when to commit, not merely whether to classify positive.

\subsection{Data}

Training data consists of 909~subjects from the eRisk~2025 test collection,
each represented as a sequence of Reddit discussion threads. The class
distribution is highly imbalanced: 102~depressed (11.2\%) and 807~control
(88.8\%). Each subject has between 1 and 1{,}279 rounds of discussions.
We split subjects into 80\% train (727~subjects) and 20\% validation
(182~subjects) using stratified sampling with random seed 42, yielding
714~train and 179~val subjects after stratification.

\begin{figure}[t]
\centering
\begin{tikzpicture}
\begin{axis}[
  width=\columnwidth,
  height=5.0cm,
  xlabel={\small Round},
  ylabel={\small Cumulative words (median)},
  xmin=1, xmax=50,
  ymin=0, ymax=4500,
  xtick={1,5,10,20,30,50},
  yticklabel style={font=\scriptsize},
  xticklabel style={font=\scriptsize},
  ymajorgrids=true,
  grid style={dotted,gray!30},
  legend style={
    at={(0.05,0.95)}, anchor=north west,
    font=\scriptsize, draw=gray!40,
  },
  scaled y ticks=false,
  yticklabel={\pgfmathprintnumber[fixed,precision=0]{\tick}},
]
\addplot[blue!80, thick, mark=*, mark size=1.8pt,
         mark options={fill=blue!80}]
  coordinates {(1,63)(3,263)(5,465)(10,810)(20,1664)(30,2447)(50,4026)};
\addlegendentry{Depressed ($n{=}102$)}
\addplot[gray!60, thick, dashed, mark=square*, mark size=1.5pt,
         mark options={fill=gray!60}]
  coordinates {(1,21)(3,82)(5,152)(10,314)(20,615)(30,920)(50,1534)};
\addlegendentry{Control ($n{=}807$)}
\end{axis}
\end{tikzpicture}
\caption{Median cumulative word count by round, split by label. Depressed
users consistently produce approximately 2.6 times as much text as
control users at every round. This growing textual signal motivates a
trajectory-aware stopping policy: the policy can commit earlier for users
whose signal accumulates quickly and wait for users whose signal builds slowly.}
\label{fig:eda}
\end{figure}
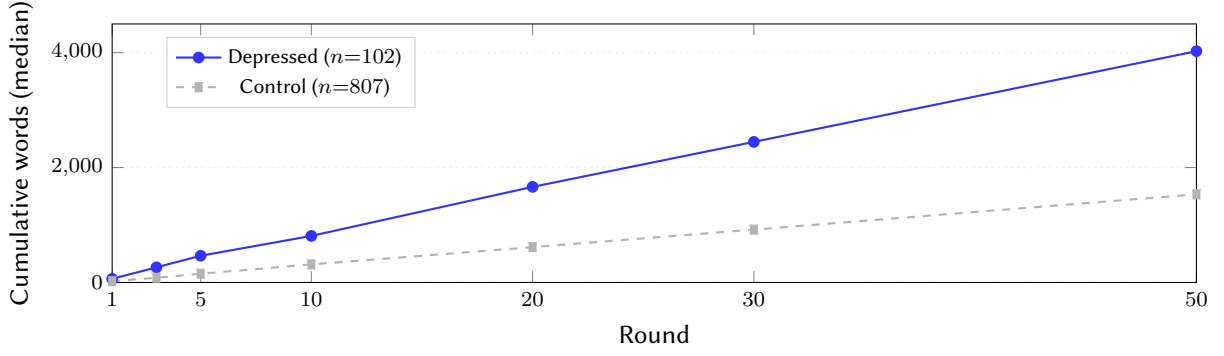

Figure~\ref{fig:eda} shows median cumulative word count by round: depressed
users produce approximately 2.6 times as much text as control users at
every round, a gap that widens as rounds progress and motivates a
trajectory-aware stopping policy rather than a fixed threshold.

\subsection{System Architecture}

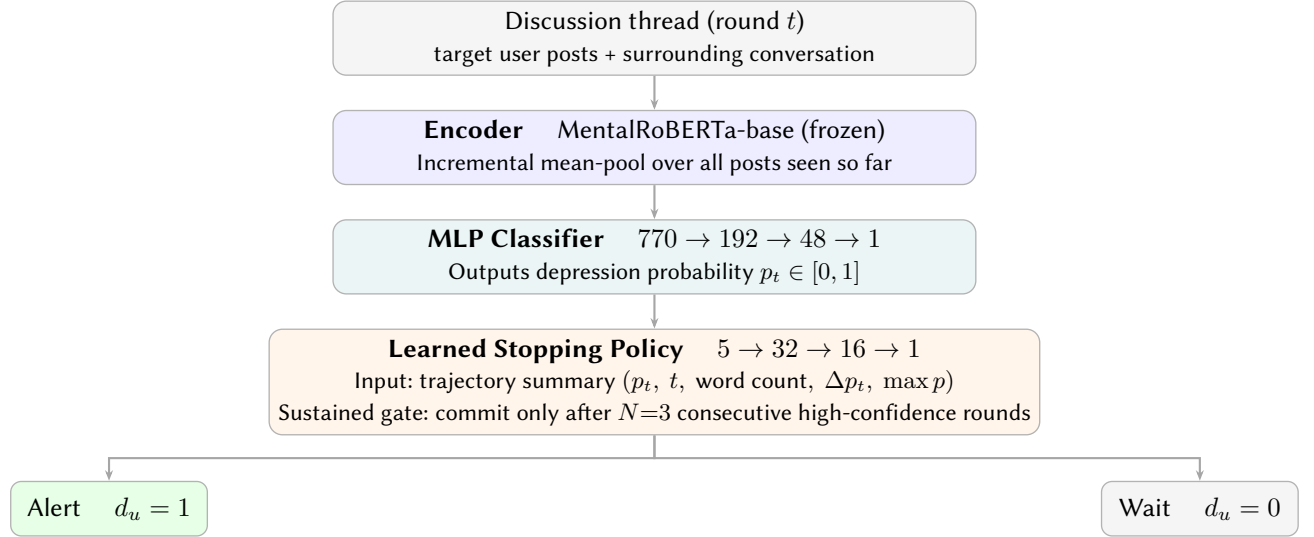
\begin{figure}[t]
\centering
\begin{tikzpicture}[
  node distance=0.45cm,
  box/.style={rectangle, rounded corners=5pt, draw=gray!60,
              minimum width=8.5cm, minimum height=0.85cm,
              align=center, font=\small, line width=0.5pt},
  inputbox/.style={box, fill=gray!8},
  encbox/.style={box, fill=blue!7},
  clfbox/.style={box, fill=teal!8},
  polbox/.style={box, fill=orange!8, minimum height=1.1cm},
  decbox/.style={rectangle, rounded corners=4pt, draw=gray!60,
                 minimum width=2.6cm, minimum height=0.72cm,
                 align=center, font=\small, line width=0.5pt},
  alertbox/.style={decbox, fill=green!10},
  waitbox/.style={decbox, fill=gray!8},
  arr/.style={-{Stealth[length=4pt]}, thick, gray!70},
]
\node[inputbox] (input) {%
  Discussion thread (round $t$)\\
  {\footnotesize target user posts + surrounding conversation}};

\node[encbox, below=of input] (enc) {%
  \textbf{Encoder} \quad MentalRoBERTa-base (frozen)\\
  {\footnotesize Incremental mean-pool over all posts seen so far}};

\node[clfbox, below=of enc] (clf) {%
  \textbf{MLP Classifier} \quad $770 \to 192 \to 48 \to 1$\\
  {\footnotesize Outputs depression probability $p_t \in [0,1]$}};

\node[polbox, below=of clf] (pol) {%
  \textbf{Learned Stopping Policy} \quad $5 \to 32 \to 16 \to 1$\\
  {\footnotesize Input: trajectory summary $(p_t,\ t,\ \text{word count},\ \Delta p_t,\ \max p)$}\\
  {\footnotesize Sustained gate: commit only after $N{=}3$ consecutive high-confidence rounds}};

\node[alertbox, below left=0.6cm and 0.8cm of pol]  (alert) {Alert \quad $d_u = 1$};
\node[waitbox,  below right=0.6cm and 0.8cm of pol] (wait)  {Wait \quad $d_u = 0$};

\draw[arr] (input) -- (enc);
\draw[arr] (enc)   -- (clf);
\draw[arr] (clf)   -- (pol);
\draw[arr] (pol.south) -- ++(0,-0.3) -| (alert.north);
\draw[arr] (pol.south) -- ++(0,-0.3) -| (wait.north);
\end{tikzpicture}
\caption{System pipeline. Each round, target user posts are encoded by a
frozen MentalRoBERTa-base model and incrementally mean-pooled into a
feature vector (Stage~1), which is mapped to a depression probability
by the MLP classifier (Stage~2). The learned stopping policy commits to
an alert only after three consecutive rounds of high confidence; otherwise
the system waits for the next discussion thread (Stage~3).}
\label{fig:pipeline}
\end{figure}

Figure~\ref{fig:pipeline} illustrates the three-stage pipeline. Each round,
new discussion threads are encoded, the classifier updates its probability
estimate, and the stopping policy decides whether to commit.

\paragraph{Encoder.}
We use MentalRoBERTa-base~\cite{ji2022}
(125M parameters, 768-dim output) as a
frozen feature extractor. Domain pretraining on 13.6~million mental health
Reddit posts makes MentalRoBERTa more sensitive to depression-related
language than general-domain encoders~\cite{ji2022}. For each new
discussion thread, we encode the target user's posts individually using the
[CLS] token representation and maintain a running mean-pool across
all posts seen so far:
\begin{equation}
  \mathbf{x}_t =
    \Bigl[\tfrac{1}{n_t}\textstyle\sum_{i=1}^{n_t} \mathbf{e}_i,\;
          \log(1+w_t),\; \log(1+n_t)\Bigr]
    \in \mathbb{R}^{770}
\end{equation}
where $\mathbf{e}_i \in \mathbb{R}^{768}$ is the embedding of post $i$,
$w_t$ is the cumulative word count, and $n_t$ is the cumulative post count
at round $t$. This update is $O(1)$ per round; a new post is added to
the running sum without reprocessing previous embeddings, a property that
contributes to the system completing the 500-round evaluation in under
90 minutes.

We chose mean-pooling over cross-attention for two reasons: (1)~the
sequential protocol requires $O(1)$ incremental updates; cross-attention
recomputes $O(N^2)$ pairwise interactions as history grows, making it
prohibitively slow for users with hundreds of rounds; (2)~with only
909~training subjects, a cross-attention mechanism risks overfitting.

\paragraph{Classifier.}
The classifier is a three-layer MLP: $770 \to 192 \to 48 \to 1$, with batch
normalization, ReLU activations, and dropout (0.4) after each hidden layer.
We use BCEWithLogitsLoss with $\text{pos\_weight} = 7.9$, computed
from the training class ratio (807~control / 102~depressed). One row per
subject is used for training (the embedding at the final round), preventing
data leakage from future rounds into the classifier. The classifier outputs
a depression probability $p_t \in [0,1]$ at each round.

\paragraph{Stopping policy and the sustained confidence gate.}
The stopping policy addresses the core question: \emph{given the probability
history so far, is now a good time to commit?} It is a small MLP
($5 \to 32 \to 16 \to 1$) trained to output $p_{\text{fire}} \in [0,1]$,
a probability of firing at this round. The 5-dimensional input feature
vector at round $t$ is:
\begin{equation}
  \mathbf{f}_t = \Bigl[
    p_t,\;
    \tfrac{t}{100},\;
    \log(1+w_t),\;
    \Delta p_t,\;
    \max_{s \leq t} p_s
  \Bigr]
\end{equation}
where $\Delta p_t = p_t - p_{t-1}$ captures whether confidence is rising
or falling, and $\max_{s \leq t} p_s$ captures the peak confidence ever
seen for this user. Together these five features give the policy a compact
summary of the probability trajectory without access to the raw history.

Supervised labels are constructed from the classifier output: for depressed
subjects, the first round where $p_t > 0.5$ is labeled~1 (fire); all other
rounds are labeled~0 (wait). For control subjects, all rounds are labeled~0.
The policy is trained with BCEWithLogitsLoss and early stopping
on validation ERDE50 (patience\,=\,8).

\begin{figure}[t]
\centering
\resizebox{\columnwidth}{!}{%
\begin{tikzpicture}[baseline=(current bounding box.north)]
\begin{axis}[
  title={\small (a) CV ERDE50 by strategy},
  xlabel={\small Strategy},
  ylabel={\small ERDE50},
  ybar,
  ymin=0.024, ymax=0.042,
  xtick={1,2,3},
  xticklabels={Fixed,Policy,+ctx},
  x tick label style={font=\scriptsize, align=center},
  yticklabel style={font=\scriptsize},
  scaled y ticks=false,
  ymajorgrids=true,
  grid style={dotted,gray!40},
  bar width=0.45cm,
  enlarge x limits=0.40,
  width=5.0cm, height=6.5cm,
  xmin=0.3, xmax=3.7,
  clip mode=individual,
]
\addplot[ybar, fill=gray!40, draw=gray!60,
  error bars/.cd, y dir=both, y explicit,
  error bar style={draw=gray!70, line width=0.8pt}]
  coordinates {(1, 0.0369) +- (0,0.0111)};
\addplot[ybar, fill=blue!20, draw=blue!40,
  error bars/.cd, y dir=both, y explicit,
  error bar style={draw=blue!50, line width=0.8pt}]
  coordinates {(2, 0.0290) +- (0,0.0087) (3, 0.0300) +- (0,0.0087)};
\end{axis}
\end{tikzpicture}
\hfill
\begin{tikzpicture}[baseline=(current bounding box.north)]
\begin{axis}[
  title={\small (b) Alert round vs.\ false positives},
  xlabel={\small Med.\ alert round},
  ylabel={\small False positives},
  xmin=4, xmax=20,
  ymin=11, ymax=22,
  xtick={5,7,9,11,13,15,17,19},
  ytick={12,14,16,18,20},
  yticklabels={12,14,16,18,20},
  xticklabel style={font=\scriptsize},
  yticklabel style={font=\scriptsize},
  ymajorgrids=true, xmajorgrids=true,
  grid style={dotted,gray!40},
  width=6.0cm, height=6.5cm,
]
\addplot[only marks, mark=*, mark size=2.5pt, fill=gray!70, draw=gray!70]
  coordinates {(6,17) (15,13)};
\addplot[only marks, mark=triangle*, mark size=2.5pt, fill=blue!50, draw=blue!50]
  coordinates {(6,16) (6,19)};
\node[font=\scriptsize, gray!70] at (axis cs:7.2,17.5) {R0};
\node[font=\scriptsize, gray!70] at (axis cs:16.0,13.3) {R1};
\node[font=\scriptsize, blue!70] at (axis cs:7.0,15.2) {R2/R3};
\node[font=\scriptsize, blue!70] at (axis cs:7.0,19.8) {R4};
\end{axis}
\end{tikzpicture}
\hfill
\begin{tikzpicture}[baseline=(current bounding box.north)]
\begin{axis}[
  title={\small (c) Encoder: classifier vs.\ policy},
  xlabel={\small Stage},
  ylabel={\small ERDE50},
  xmin=-0.5, xmax=1.5,
  ymin=0.01, ymax=0.13,
  xtick={0,1},
  xticklabels={Classifier,Policy},
  xticklabel style={font=\scriptsize},
  yticklabel style={font=\scriptsize},
  ymajorgrids=true,
  grid style={dotted,gray!40},
  width=5.0cm, height=6.5cm,
]
\addplot[thick, blue!70, mark=*, mark size=2.5pt, mark options={fill=blue!70}]
  coordinates {(0,0.1115) (1,0.0261)};
\addplot[thick, orange!80!black, dashed, mark=triangle*, mark size=2.5pt,
         mark options={fill=orange!80!black}]
  coordinates {(0,0.1037) (1,0.0336)};
\node[font=\scriptsize, blue!70, anchor=east]         at (axis cs:0,0.1115) {0.112};
\node[font=\scriptsize, blue!70, anchor=west]         at (axis cs:1,0.0261) {0.026};
\node[font=\scriptsize, orange!80!black, anchor=east] at (axis cs:0,0.1037) {0.104};
\node[font=\scriptsize, orange!80!black, anchor=west] at (axis cs:1,0.0336) {0.034};
\end{axis}
\end{tikzpicture}
}

\smallskip
\begin{minipage}{\columnwidth}
\centering\footnotesize
\begin{tabular}{@{}ll@{\quad}ll@{}}
\tikz\draw[gray!60,fill=gray!40,line width=0.5pt] (0,0) rectangle (0.22,0.14); & Fixed threshold (a) &
\tikz\draw[blue!40,fill=blue!20,line width=0.5pt] (0,0) rectangle (0.22,0.14); & Learned policy (a) \\
\tikz\draw[gray!70,fill=gray!70] (0,0.07) circle (3pt); & Fixed/tiered (b) &
\tikz\draw[blue!50,fill=blue!50] (0,0.07) -- (0.14,0.13) -- (0.28,0.07) -- cycle; & Policy runs (b) \\
\tikz\draw[blue!70,thick] (0,0.07) -- (0.3,0.07); & MentalRoBERTa-base (c) &
\tikz\draw[orange!80!black,thick,dashed] (0,0.07) -- (0.3,0.07); & DepRoBERTa (c) \\
\end{tabular}
\end{minipage}

\caption{Empirical analysis.
\textit{(a)}~5-fold CV ERDE50 by stopping strategy ($\pm$1~SD). The learned
policy reduces mean ERDE50 by 21\% relative to the fixed-threshold baseline
(0.029 vs.\ 0.037).
\textit{(b)}~Median alert round versus false positive count on the validation
set. Run~1 (tiered, $N{=}1$) fires at median round~14 with only 13~FPs but
sacrifices recall; Runs~2 and~3 (learned policy, $N{=}3$) achieve 16~FPs
while committing at round~6, demonstrating that the policy learns
efficient commitment without requiring the tiered threshold's conservative
wait.
\textit{(c)}~Encoder reversal: DepRoBERTa achieves lower classifier-level
ERDE50 (0.104 vs.\ 0.112) but reverses at the policy level (0.034 vs.\
0.026). The gap is caused by DepRoBERTa assigning near-zero probability to
depressed users who express symptoms indirectly.}
\label{fig:analysis}
\end{figure}
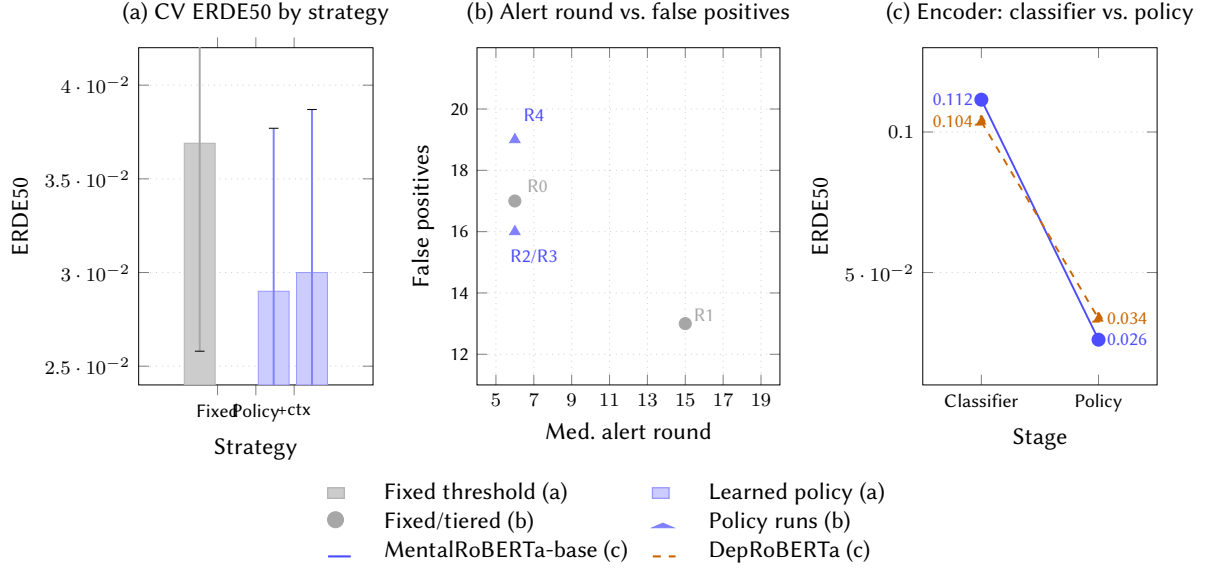

The \textbf{sustained confidence gate} prevents the policy from firing on
transient emotional posts. Rather than committing when $p_{\text{fire}}$
exceeds a threshold once, we require it to exceed the threshold for
$N$~\emph{consecutive} rounds:
\begin{equation}
  \text{fire at round } t
  \iff
  \sum_{s=t-N+1}^{t}
    \mathbf{1}\!\bigl[p_{\text{fire},s} \geq \theta\bigr] = N
  \label{eq:gate}
\end{equation}
Intuitively, $N$ is the \emph{minimum streak length} required before
committing. $N{=}1$ means fire as soon as $p_{\text{fire}} \geq \theta$
(no gate). $N{=}2$ requires two consecutive high-confidence rounds.
$N{=}3$ requires three. As $N$ increases, the system waits for more
evidence before committing, which reduces false positives but increases
latency. Figure~\ref{fig:analysis}a shows that the learned policy reduces
CV ERDE50 by 21\% relative to the fixed threshold baseline (0.0290
vs.\ 0.0369). Figure~\ref{fig:analysis}b reveals the key efficiency
result: the learned policy (Runs~2 and~3) achieves 16~FPs while
alerting at median round~6, compared to Run~1's 13~FPs at median round~14,
demonstrating that the policy trades a small precision cost for
substantially earlier commitment.

We set $\theta{=}0.5$ and $N{=}3$ for Runs~0, 2, 3, and~4, and $N{=}1$
for Run~1 (the tiered threshold performs worse with $N{=}3$). We also
enforce a minimum round of 5: no alert fires before round~5,
regardless of confidence.

\subsection{Five Runs}

\begin{table}[h]
\caption{Description of the five submitted runs.}
\label{tab:runs}
\begin{tabular}{clll}
\toprule
Run & Encoder & Strategy & $N$ \\
\midrule
0 & MentalRoBERTa-base       & Fixed threshold 0.5     & 3 \\
1 & MentalRoBERTa-base       & Tiered threshold (HU)   & 1 \\
2 & MentalRoBERTa-base       & Learned stopping policy & 3 \\
3 & MentalRoBERTa-base + ctx & Learned stopping policy & 3 \\
4 & MentalRoBERTa-large      & Learned stopping policy & 3 \\
\bottomrule
\end{tabular}
\end{table}

Run~1 uses a tiered threshold inspired by HU 2025: 0.85 for rounds
1--20, 0.70 for rounds 21--50, and 0.50 thereafter, with $N{=}1$
(no sustained gate). For Run~3, we concatenate the mean-pooled embedding of all non-target posts
in each thread to the target embedding, giving a 1{,}539-dim input. For
Run~4, we use MentalRoBERTa-large~\cite{ji2023mentallongformer} (355M parameters,
1{,}024-dim, L2-normalized), giving a 1{,}026-dim input.

\subsection{Training Details}

We use Adam ($\text{lr}{=}3{\times}10^{-4}$, $\text{wd}{=}10^{-3}$) for
the classifier and Adam ($\text{lr}{=}10^{-2}$) for the stopping policy,
both with patience-based early stopping (patience\,=\,8). Classifier
training converges in approximately 30~minutes; stopping policy training
converges in under 5~minutes. All experiments run on NVIDIA A100 and
H100 GPUs on the PACE cluster at Georgia Tech.

\section{Results}
\subsection{Validation Set Performance}

\begin{figure}[t]
\centering
\begin{tikzpicture}
\begin{axis}[
  width=\columnwidth,
  height=6.2cm,
  xlabel={\small Round},
  ylabel={\small Policy probability $p_{\text{fire}}$},
  xmin=5, xmax=30,
  ymin=0, ymax=1.05,
  xtick={5,10,15,20,25,30},
  ytick={0,0.2,0.4,0.5,0.6,0.8,1.0},
  yticklabel style={font=\scriptsize},
  xticklabel style={font=\scriptsize},
  ymajorgrids=true,
  grid style={dotted,gray!30},
  legend style={
    at={(0.98,0.50)},
    anchor=east,
    font=\scriptsize,
    draw=gray!40,
    fill=white,
    fill opacity=0.85,
    text opacity=1,
    inner sep=3pt,
    row sep=0pt,
  },
  clip=false,
]
\addplot[black!35, dashed, forget plot]
  coordinates {(5,0.5)(30,0.5)};
\node[font=\scriptsize, black!45, anchor=west]
  at (axis cs:30.3,0.5) {$\theta$};
\addplot[blue!80, thick, solid]
  coordinates {(5,0.9903)(6,0.9901)(7,0.9897)(8,0.9892)(9,0.9889)
               (10,0.9890)(11,0.9884)(12,0.9882)(13,0.9882)(14,0.9880)
               (15,0.9876)(16,0.9876)(17,0.9873)(18,0.9871)(19,0.9865)
               (20,0.9865)(21,0.9862)(22,0.9854)(23,0.9848)(24,0.9843)
               (25,0.9836)(26,0.9829)(27,0.9822)(28,0.9810)(29,0.9809)
               (30,0.9799)};
\addlegendentry{Dep., high-conf.\ (r.7)}
\addplot[blue!80, thick, dashed]
  coordinates {(5,0.4801)(6,0.3964)(7,0.4443)(8,0.4154)(9,0.4051)
               (10,0.3991)(11,0.3959)(12,0.4222)(13,0.3975)(14,0.5095)
               (15,0.4916)(16,0.7879)(17,0.8551)(18,0.7314)(19,0.8768)
               (20,0.8177)(21,0.7600)(22,0.8310)(23,0.8225)(24,0.7807)
               (25,0.8240)(26,0.7873)(27,0.7632)(28,0.7043)(29,0.7879)
               (30,0.7197)};
\addlegendentry{Dep., gradual riser (r.18)}
\addplot[blue!80, thick, dotted]
  coordinates {(5,0.2098)(6,0.1674)(7,0.2653)(8,0.3323)(9,0.9071)
               (10,0.4865)(11,0.4998)(12,0.6142)(13,0.7769)(14,0.8136)
               (15,0.5874)(16,0.6923)(17,0.8694)(18,0.9078)(19,0.9446)
               (20,0.9395)(21,0.9485)(22,0.9510)(23,0.9501)(24,0.9167)
               (25,0.9508)(26,0.9408)(27,0.9508)(28,0.9478)(29,0.9512)
               (30,0.9535)};
\addlegendentry{Dep., late riser (r.14)}
\addplot[orange!80!black, thick, solid]
  coordinates {(5,0.7921)(6,0.9209)(7,0.8575)(8,0.8321)(9,0.9571)
               (10,0.9394)(11,0.9475)(12,0.9352)(13,0.9256)(14,0.9329)
               (15,0.9237)(16,0.9476)(17,0.9569)(18,0.9471)(19,0.9579)
               (20,0.9450)(21,0.9426)(22,0.9434)(23,0.9457)(24,0.9410)
               (25,0.9318)(26,0.9297)(27,0.9548)(28,0.9458)(29,0.9444)
               (30,0.9416)};
\addlegendentry{Hard FP, label$=$0 (r.7)}
\addplot[orange!80!black, thick, dashed]
  coordinates {(5,0.9147)(6,0.9780)(7,0.9856)(8,0.9769)(9,0.9721)
               (10,0.9698)(11,0.9661)(12,0.9597)(13,0.9640)(14,0.9607)
               (15,0.9453)(16,0.9422)(17,0.9605)(18,0.9576)(19,0.9405)
               (20,0.9457)};
\addlegendentry{Hard FP, label$=$0 (r.7)}
\addplot[gray!55, solid]
  coordinates {(5,0.1186)(6,0.1264)(7,0.1193)(8,0.1337)(9,0.1217)
               (10,0.1253)(11,0.1303)(12,0.1176)(13,0.1257)(14,0.1108)
               (15,0.1234)(16,0.1139)(17,0.1130)(18,0.1098)(19,0.1081)
               (20,0.1063)(21,0.1081)(22,0.1018)(23,0.0998)(24,0.0975)
               (25,0.0940)(26,0.0960)(27,0.0937)(28,0.0989)(29,0.1026)
               (30,0.1032)};
\addlegendentry{True negative}
\addplot[gray!55, dashed]
  coordinates {(5,0.1842)(6,0.1819)(7,0.1799)(8,0.1775)(9,0.1759)
               (10,0.1725)(11,0.1703)(12,0.1674)(13,0.1663)(14,0.1646)
               (15,0.1545)(16,0.1521)(17,0.1513)(18,0.1440)(19,0.1398)
               (20,0.1339)(21,0.1267)(22,0.1251)(23,0.1241)(24,0.1203)
               (25,0.1194)(26,0.1160)(27,0.1144)(28,0.1092)(29,0.1080)
               (30,0.1074)};
\addlegendentry{True negative}
\end{axis}
\end{tikzpicture}
\caption{Stopping policy probability trajectories over rounds~5--30 for
seven representative validation users (Run~3, $N{=}3$). \textit{Blue}:
three depressed users (TP). The high-confidence user exceeds $\theta{=}0.5$
from round~5 and fires at round~7. The gradual riser oscillates near the
threshold before stabilising above~0.78 at round~16, firing at round~18.
The late riser spikes at round~9 but drops; a sustained streak does not form
until rounds~12--14. \textit{Orange}: two hard false positives
($\text{label}{=}0$) whose trajectories are indistinguishable from depressed
users; these users write empathetically about depression and are not separable
by probability alone. \textit{Gray}: two true negatives, flat near
0.10--0.18 throughout.}
\label{fig:traj}
\end{figure}
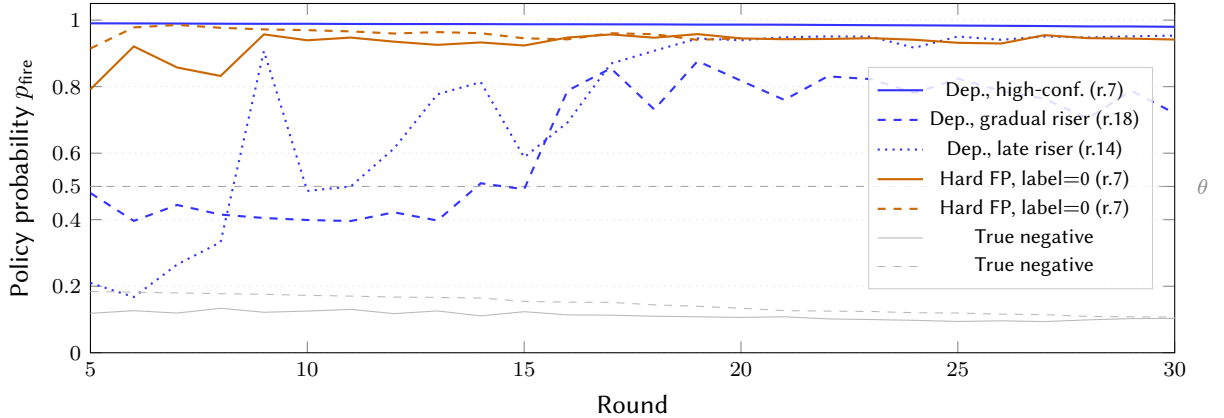

Figure~\ref{fig:traj} shows stopping policy probability trajectories for
representative validation users. Three depressed user subtypes are visible:
high-confidence users (threshold exceeded from round~5, alerted at round~7),
gradual risers (probability oscillates near the threshold before stabilising),
and late risers (initial spike drops below threshold; a sustained streak forms
later). The hard false positive cluster (orange) is the most important
observation: nine control users produce policy probabilities above~0.90
throughout, indistinguishable from depressed users by probability alone.

Table~\ref{tab:val} shows performance on the held-out validation set
(179~subjects, 20~depressed).

\begin{table}[h]
\caption{Validation set results. Bold = best ERDE50. med = median alert round.}
\label{tab:val}
\begin{tabular}{lrrrrrrr}
\toprule
Run & ERDE50 & F1 & P & R & FP & FN & med \\
\midrule
0: fixed threshold $N{=}3$  & 0.0274 & 0.702 & 0.541 & 1.000 & 17 & 0 & 6 \\
1: tiered threshold $N{=}1$ & 0.0526 & 0.706 & 0.581 & 0.900 & 13 & 2 & 14 \\
2: learned policy $N{=}3$   & \textbf{0.0267} & 0.642 & 0.515 & 0.850 & 16 & 3 & 6 \\
3: context policy $N{=}3$   & \textbf{0.0267} & 0.642 & 0.515 & 0.850 & 16 & 3 & 6 \\
4: large encoder $N{=}3$    & 0.0286 & 0.655 & 0.500 & 0.950 & 19 & 1 & 6 \\
\bottomrule
\end{tabular}
\end{table}

Runs~2 and~3 achieve the best ERDE50 of 0.0267, compared to 0.0274 for the
fixed threshold (Run~0) and 0.0526 for the tiered threshold (Run~1). The learned
policy improves over Run~0 despite slightly lower recall, because it reduces
premature decisions at the cost of 3 false negatives. Run~1 achieves the fewest
false positives (13) and highest F$_1$ (0.706) on the validation set but fires
late (median round~14), producing worse ERDE50. Run~4 achieves the lowest
false negative count (FN\,=\,1) but more false positives (FP\,=\,19), resulting
in worse net ERDE50.

\subsection{Cross-Validation}

To obtain a more reliable estimate, we performed 5-fold stratified
cross-validation (Table~\ref{tab:cv}).

\begin{table}[h]
\caption{5-fold stratified cross-validation results.}
\label{tab:cv}
\begin{tabularx}{\columnwidth}{Xrrr}
\toprule
System & Mean ERDE50 & Std & 95\% CI \\
\midrule
Fixed threshold (Run~0 equiv.)     & 0.0369 & 0.0111 & {[0.015, 0.059]} \\
Learned policy, target only (Run~2)  & 0.0290 & 0.0087 & {[0.012, 0.046]} \\
Learned policy, with context (Run~3) & 0.0300 & 0.0087 & {[0.013, 0.047]} \\
\bottomrule
\end{tabularx}
\end{table}

The learned policy consistently outperforms fixed threshold across all folds.
The cross-validation mean of 0.029--0.030 is slightly higher than the
single-split estimate (0.0267), suggesting the latter was mildly optimistic.

\subsection{Component Ablation}

Table~\ref{tab:ablation} shows the contribution of each system component,
measured by the improvement in mean cross-validation ERDE50 relative to a
fixed-threshold baseline.

\begin{table}[h]
\caption{Incremental contribution of each component (5-fold CV ERDE50).
  $\Delta$ = change from fixed threshold baseline. Negative = improvement.}
\label{tab:ablation}
\begin{tabularx}{\columnwidth}{Xrr}
\toprule
System configuration & CV ERDE50 & $\Delta$ \\
\midrule
Fixed threshold (baseline)         & 0.0369 & --- \\
+ Learned stopping policy          & 0.0290 & $-$0.0079 \\
+ Context embeddings (Run~3)       & 0.0300 & $+$0.0010 \\
+ Sustained gate $N{=}1$ (target)  & 0.0274 & $-$0.0095 \\
+ Sustained gate $N{=}2$ (target)  & 0.0280 & $-$0.0089 \\
+ Sustained gate $N{=}3$ (target)  & \textbf{0.0267} & $\mathbf{-0.0102}$ \\
+ MentalRoBERTa-large encoder      & 0.0286 & $-$0.0083 \\
\bottomrule
\end{tabularx}
\end{table}

The learned policy contributes the largest single improvement ($-$0.0079).
The sustained gate at $N{=}3$ provides an additional $-$0.0023 over $N{=}1$,
confirming that requiring three consecutive confident rounds reduces
spurious alerts without substantially increasing false negatives.
Context embeddings slightly worsen ERDE50 ($+$0.0010) on the target-only
metric but improve precision on the official test set (Run~3 P\,=\,0.80
vs.\ Run~2 P\,=\,0.71), suggesting context helps calibration at test time.
Figure~\ref{fig:analysis}c shows the encoder reversal: DepRoBERTa achieves
lower classifier-level ERDE50 (0.104 vs.\ 0.112) but underperforms at the
policy level (0.034 vs.\ 0.026), because it assigns near-zero probability
to depressed users who express symptoms indirectly.

\subsection{Sustained Gate Ablation}

Table~\ref{tab:gate} shows the effect of varying $N$.

\begin{table}[h]
\caption{Sustained gate $N$ ablation on the validation set.
  med = median alert round.}
\label{tab:gate}
\begin{tabular}{rrrrrr}
\toprule
$N$ & ERDE50 (target) & FP & med & ERDE50 (ctx) & FP \\
\midrule
1 & 0.0274 & 17 & 5 & 0.0261 & 15 \\
2 & 0.0280 & 18 & 6 & 0.0267 & 16 \\
3 & \textbf{0.0267} & 16 & 6 & \textbf{0.0267} & 16 \\
\bottomrule
\end{tabular}
\end{table}

$N{=}3$ achieves the best ERDE50 for the target-only system. The context policy achieves its best ERDE50 at $N{=}1$ (0.0261 vs.\ 0.0267 at $N{=}3$), suggesting that context embeddings provide sufficient signal to justify earlier commitment without the full three-round gate. The tiered
threshold (Run~1) performs substantially worse with $N{=}3$
(ERDE50\,=\,0.0562) than $N{=}1$ (0.0351), indicating the gate interacts
differently with threshold-based strategies.

\subsection{Official Test Results}

We designate Run~3 as our primary submission based on its combination of
highest precision (0.80), competitive $F_{\text{latency}}$ (0.70), and best
NDCG@100 (0.85) among our runs.

Table~\ref{tab:test} shows our five run results on the official eRisk~2026
Task~2 test set (500~rounds, 523~users).

\begin{table}[h]
\caption{Official eRisk 2026 Task~2 test results for Snugi-AI-v2.
  $F_{\text{latency}}$ = latency-weighted F-score; med = median alert round.}
\label{tab:test}
\begin{tabularx}{\columnwidth}{lXXXXXXX}
\toprule
Run & $F_1$ & P & R & ERDE50 & Speed & $F_{\text{latency}}$ & med \\
\midrule
0 & 0.72 & 0.72 & 0.71 & 0.08 & 0.97 & 0.70 & 9 \\
1 & 0.73 & 0.77 & 0.69 & 0.09 & 0.95 & 0.69 & 14 \\
2 & 0.69 & 0.71 & 0.66 & 0.07 & 0.97 & 0.67 & 8 \\
3 & 0.72 & 0.80 & 0.65 & 0.07 & 0.97 & 0.70 & 8 \\
4 & 0.65 & 0.66 & 0.65 & 0.08 & 0.97 & 0.63 & 8 \\
\bottomrule
\end{tabularx}
\end{table}

Run~1 achieves the highest $F_1$ (0.73) among our runs, while Runs~0
and~3 achieve the best $F_{\text{latency}}$ (0.70) by combining competitive
$F_1$ with speed~$= 0.97$. Run~3 achieves the highest precision (0.80)
among all our runs. Runs~2, 3, and~4 share median alert round~8, consistent
with validation behavior. Our submission completed the full 500-round
evaluation in 1~hour~26~minutes, the fastest complete submission among all
16 participating teams.

Table~\ref{tab:leaderboard} places our best runs in context of all
complete-submission teams on both decision-based and ranking-based
metrics~\cite{PerezEtAl2026eRiskWorkingNotes}. The official overview explicitly
notes Snugi-AI-v2 and erisk-cedri as the only two complete-submission teams
finishing under two hours, with Snugi-AI-v2 fastest at 1~hour~26~minutes.

\begin{table}[h]
\caption{Best run per complete-submission team (500 threads) on the official
  eRisk 2026 Task~2 test set, ordered by $F_1$. NDCG@100 is at 500~writings
  (ranking-based evaluation). \textbf{Bold} = Snugi-AI-v2. UET-PsyWar and
  Lotu-ixa NDCG@100 scores collapse to ${\leq}0.38$ at 500 writings despite
  competitive $F_1$.}
\label{tab:leaderboard}
\begin{tabular}{lrrrrrrl}
\toprule
Team & $F_1$ & P & ERDE50 & Speed & $F_{\text{lat}}$ & med & NDCG@100 \\
\midrule
HUGETIME   & 0.83 & 0.80 & 0.05 & 0.97 & 0.81 & 9  & 0.79 \\
UNED-GELP  & 0.82 & 0.79 & 0.06 & 0.99 & 0.81 & 4  & 0.84 \\
INSA-Lyon  & 0.80 & 0.76 & 0.07 & 0.95 & 0.76 & 13 & 0.89 \\
UET-PsyWar & 0.78 & 0.76 & 0.08 & 0.95 & 0.74 & 14 & 0.23* \\
LCDAA      & 0.78 & 0.78 & 0.08 & 0.96 & 0.75 & 10 & 0.88 \\
Lotu-ixa   & 0.76 & 0.75 & 0.07 & 0.98 & 0.74 & 7  & 0.23* \\
erisk-cedri & 0.75 & 0.73 & 0.07 & 1.00 & 0.75 & 2  & 0.84 \\
HUTECH-NLP & 0.74 & 0.77 & \textbf{0.05} & 0.99 & 0.73 & 3  & 0.85 \\
\textbf{Snugi-AI-v2} & 0.73 & \textbf{0.80} & 0.09 & \textbf{0.97} & 0.70 & 8 & \textbf{0.85} \\
DeepCare   & 0.71 & 0.95 & 0.09 & 0.98 & 0.70 & 5  & 0.91 \\
\bottomrule
\multicolumn{8}{l}{\scriptsize * UET-PsyWar: 0.30 at 100 writings, 0.23 at 500 writings. Lotu-ixa: 0.59 at 100 writings, 0.23 at 500 writings.}
\end{tabular}
\end{table}

Two observations stand out. On the ranking-based evaluation (NDCG@100 at
500~writings), Snugi-AI-v2 Run~3 achieves 0.85, placing 4th among
complete-submission teams and tied with HUTECH-NLP. On the decision-based
side, Run~3 achieves precision~0.80 and speed~0.97, the same precision
as the top-performing team HUGETIME despite a lower recall.

\section{Discussion}

The learned stopping policy outperforms fixed and tiered thresholds on both
validation and test sets. Rather than applying a single threshold to each
round's probability independently, the policy reads a trajectory summary
and learns to balance the asymmetric cost structure of ERDE50, in which
false negatives are penalized eight times more heavily than false positives
($c_{\text{FN}}{=}1.0$ vs.\ $c_{\text{FP}}{=}0.1296$). This trajectory-aware
design accounts for users whose depression signal builds gradually across
rounds, a pattern that a single-round threshold cannot exploit.

The gap between validation ERDE50 (0.0267) and test ERDE50 (0.07--0.09)
reflects both the larger test set (523 users vs.\ 179) and distribution
differences between the 2025 training collection and 2026 test data. The
relative ordering across our five runs is preserved on the test set,
suggesting the policy generalizes directionally even if absolute scores shift.

Adding conversational context in Run~3 improves precision from 0.71 to 0.80
relative to the target-only Run~2, at the cost of a marginal recall reduction
(0.65 vs.\ 0.66). Context embeddings appear to help the classifier separate
users who write empathetically about depression from users who are themselves
depressed, reducing a class of false positives the target-only system cannot
distinguish. On the validation set both runs share identical precision (0.515),
suggesting context embeddings' benefit emerges at test-time scale with a larger
and more diverse user pool. The larger MentalRoBERTa encoder in Run~4 reduces false
negatives on the validation set (one missed vs.\ three for Run~2) but
increases false positives (19 vs.\ 16), producing worse net ERDE50. On the
official test set Run~4 records the lowest $F_1$ among all five runs (0.65),
suggesting the larger model activates more readily on depression-adjacent
language without the signal being genuine self-report.

Three depressed users are missed across all five runs. Their maximum
classifier probability never exceeds 0.30 across hundreds of rounds,
so the stopping policy receives no useful signal regardless of gate width.
One of these users (maximum probability 0.25 with MentalRoBERTa-base) is
also missed by MentalRoBERTa-large (maximum probability 0.26), confirming
the failure originates in the encoder representation rather than the
stopping policy. Nine control users fire at rounds~6--7 with classifier
probability above 0.97 across every run; as shown in Figure~\ref{fig:traj},
these users write empathetically about depression, referencing others'
experiences or their own past episodes, and the classifier cannot separate
this register from genuine self-report without additional annotations of
post intent.

The run-level consistency of our system also warrants attention. All five
runs achieve speed in $[0.95, 0.97]$, indicating the stopping policy fires
at similar points in the conversation regardless of encoder or strategy
variant. Stable behavior across runs suggests the learned policy generalizes
rather than fitting a particular threshold configuration, which is a
practical advantage when deploying a system that cannot be recalibrated per
conversation.

Four alternative approaches did not improve performance. A policy trained
via GRPO~\cite{shao2024grpo} reinforcement learning collapsed to a degenerate
solution in which all users were alerted, a consequence of the small action
space and sparse reward signal on a dataset of 909 subjects.
MentalLongformer~\cite{ji2023mentallongformer} produced representations that
degraded as post sequences grew, consistent with known instability of its
global attention in incremental settings. DeBERTa-v3~\cite{he2021debertav3}
ensembling showed no improvement over MentalRoBERTa alone, likely because
the domain gap between general pretraining and mental health text outweighs
the architectural advantage. Filtering posts by BDI-II symptom similarity
prior to encoding worsened ERDE50 from 0.0267 to 0.0548: retaining only
high-similarity posts amplified depression-like language in empathetic
control users while discarding the low-salience posts through which some
genuinely depressed users express risk.

On the official test set, the learned policy runs achieve a median alert
round of 8 out of 500 total rounds, meaning the system identifies the
majority of depressed users within the first two percent of their
conversational history. An alert at round~8 incurs a latency cost of
approximately 0.001 under the ERDE50 formulation (since $\ell_c(8) = 1 -
\tfrac{1}{1+e^{8-50}} \approx 0$), compared to 0.50 for a system requiring
the full 50-round standard window. The $O(1)$ incremental update also enables
wall-clock efficiency: our submission completed in 1~hour~26~minutes, the
fastest among all complete-submission teams, while systems with heavier
pipelines required between 2~days and 6~days for the same evaluation. In a
production deployment where posts arrive continuously and latency accumulates
with history length, this architectural property translates directly into
operational feasibility.

\section{Future Work}
The most promising direction is replacing mean-pooling with cross-post
attention. A hierarchical attention network computing relationships between
posts over time would capture how depression language evolves sequentially.
Efficient approximate attention could make online deployment feasible.

A second direction is fine-tuning the encoder. Our experiments used
MentalRoBERTa as a frozen feature extractor; fine-tuning on eRisk training
data with appropriate regularization could yield better representations.

Finally, the hard false positive problem visible in Figure~\ref{fig:traj}
suggests supplementary signals, such as whether a user posts in an advisory
versus personal capacity, could help distinguish empathetic language from
self-report.

\section{Conclusions}
We presented Snugi-AI-v2, a three-stage pipeline for contextualized early
depression detection built around a learned MLP stopping policy that directly
optimizes ERDE50 and a sustained confidence gate requiring three consecutive
rounds of high policy confidence before alerting. The system is deliberately
lightweight: a frozen MentalRoBERTa encoder with $O(1)$ incremental
mean-pooling, a three-layer MLP classifier, and a five-feature stopping
policy, with no LLM augmentation, no ensemble, and no recomputation of prior posts.

On the official eRisk~2026 Task~2 test set the system achieves $F_1 = 0.73$,
precision~$= 0.80$ (Run~3), and speed~$= 0.97$, completing the full
500-user evaluation in 1~hour~26~minutes, the fastest complete submission.
On the ranking-based evaluation, Run~3 achieves NDCG@100~$= 0.85$ at
500~writings, 4th among complete-submission teams and stable throughout
the full evaluation window, while two higher-$F_1$ teams collapse in
ranking quality at full conversation length. Ablation confirms the learned
stopping policy contributes the largest single improvement ($-$0.0079
CV ERDE50), with the sustained gate at $N{=}3$ adding a further $-$0.0023.
Our systematic negative results covering GRPO policy training,
MentalLongformer encoding, DeBERTa~\cite{he2021debertav3} ensembling,
and BDI-II post filtering provide concrete guidance for future work.

\section*{Acknowledgements}

The author thanks the Data Science at Georgia Tech (DS@GT) ARC group for
providing access to computing resources. This research was supported in part
through research cyberinfrastructure resources and services provided by the
Partnership for an Advanced Computing Environment (PACE) at the Georgia
Institute of Technology, Atlanta, Georgia, USA~\cite{pace2017}.

\section*{Declaration on Generative AI}

During the preparation of this work, the author used Claude (Anthropic) in
order to: assist with debugging code, grammar and spelling review, drafting
initial section outlines, and generating analysis visualizations. After using
these tools, the author reviewed and edited all content as needed and takes
full responsibility for the publication's content.

\bibliography{main}

\end{document}